\documentclass[journal,twoside]{IEEEtran}
\usepackage{microtype}
\usepackage{graphicx}
\usepackage{subfigure}
\usepackage{booktabs} 
\usepackage{amssymb,amsmath}

\usepackage{color}
\usepackage[english]{babel}
\usepackage{array}
\usepackage{bm}
\usepackage{multirow}
\usepackage{enumitem}
\usepackage{pifont}
\usepackage{tikz}
\usepackage{amssymb}
\usepackage{enumitem}
\usepackage[english]{babel}
\usepackage{array}
\usepackage{bm}
\usepackage{multirow}
\usepackage{enumitem}
\usepackage{pifont}
\usepackage{flushend}

\newcommand{\todo}[1]{\textcolor[rgb]{0.00,0.00,0.00}{#1}}

\newcommand{\minorR}[1]{\textcolor[rgb]{0.00,0.00,0.00}{#1}}

\begin{document}

\title{Hardware/Software Co-Exploration of Neural Architectures}

\author{Weiwen Jiang,
        Lei Yang,
        Edwin H.-M. Sha,~\IEEEmembership{Senior Member,~IEEE,}
        Qingfeng Zhuge, Shouzhen Gu,\\
        Sakyasingha Dasgupta,~\IEEEmembership{Member,~IEEE,}
        Yiyu Shi,~\IEEEmembership{Senior Member,~IEEE,}
        and~Jingtong Hu,~\IEEEmembership{Member,~IEEE}
\thanks{
   W. Jiang, L. Yang and Y. Shi are  with the Department of Computer Science and Engineering, University of Notre Dame, Notre Dame, IN 46556 (e-mail: wjiang2@nd.edu; lyang24@nd.edu yshi4@nd.edu).
}
\thanks{E. H.-M. Sha, Q. Zhuge, and S. Gu are with the School of Computer Science and Software Engineering, East China Normal University, 200062 China
}
\thanks{S. Dasgupta is with Edgecortix Inc., Tokyo, Japan, 1410031.
}
\thanks{J. Hu is with the Department of Electrical and Computer Engineering, University of Pittsburgh, Pittsburgh, PA 15261 (e-mail: jthu@pitt.edu).
}
}

\markboth{IEEE TRANSACTIONS ON COMPUTER-AIDED DESIGN OF INTEGRATED CIRCUITS AND SYSTEMS}%
{Jiang \MakeLowercase{\textit{et al.}}: Hardware/Software Co-Exploration of Neural Architectures}

\maketitle

\begin{abstract}

We propose a novel hardware and software co-exploration framework for efficient neural architecture search (NAS). Different from existing hardware-aware NAS which assumes a fixed hardware design and explores the \textit{neural architecture search space} only,  
our framework simultaneously explores both the architecture search space and the \textit{hardware design space} to identify the best neural architecture and hardware pairs that maximize both test accuracy and hardware efficiency. 
Such a practice greatly opens up the design freedom and pushes forward the Pareto frontier between hardware efficiency and test accuracy for better design tradeoffs. 
The framework iteratively performs a two-level (fast and slow) exploration.
Without lengthy training, the fast exploration can effectively fine-tune hyperparameters and prune inferior architectures in terms of hardware specifications, which significantly accelerates the NAS process.
Then, the slow exploration trains candidates on a validation set and updates a controller using the reinforcement learning to maximize the expected accuracy together with the hardware efficiency. 
\minorR{In this paper, we demonstrate that the co-exploration framework can effectively expand the search space to incorporate models with high accuracy, and we theoretically show that the proposed two-level optimization can efficiently prune inferior solutions to better explore the search space. Experimental results on ImageNet show that the co-exploration NAS can find solutions with the same accuracy, $35.24\%$ higher throughput, $54.05\%$ higher energy efficiency, compared with the hardware-aware NAS.}
\end{abstract}


\setlength{\textfloatsep}{6pt}
\setlength{\floatsep}{6pt}
\setlength{\dbltextfloatsep}{6pt}

\section{Introduction} \label{sec:Intro}

\begin{figure}[t]
\begin{center}
\centerline{\includegraphics[width=\columnwidth]{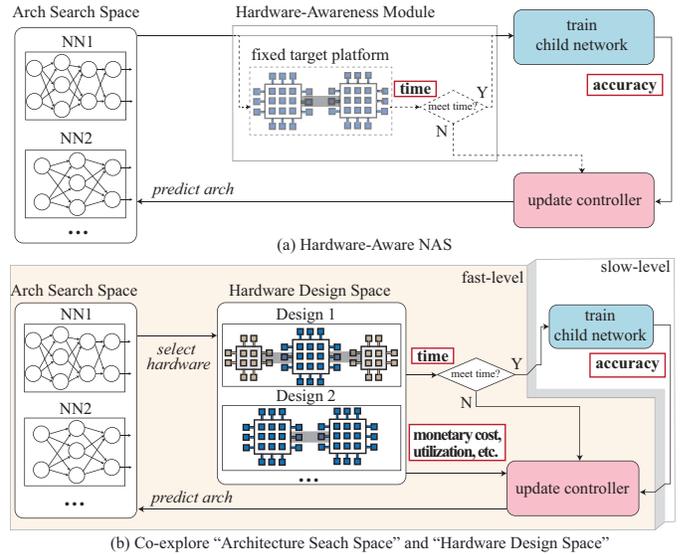}}
\caption{Comparison between (a) hardware-aware NAS; (b) the proposed  hardware/software co-exploration NAS. The red rectangles convey the metrics that can be optimized in the exploration.}\label{Fig:IntrIll}
\end{center}
\end{figure}

\todo{Neural architecture search (NAS) has achieved great success to liberate human labor in the design of neural architectures for various tasks including image classification, image segmentation and language modeling \cite{cai2018efficient,zoph2017learning,real2017large,liu2017hierarchical,nekrasov2019architecture}.} Most recently, targeting a fixed hardware platform, the hardware-aware NAS \cite{wu2018fbnet,tan2018mnasnet,cai2018proxylessnas} has been proposed to take into consideration the estimated timing performance (such as latency or throughput) in addition to accuracy (see Figure \ref{Fig:IntrIll}(a)). 

All of the existing NAS frameworks explore the \textit{architecture search space} only, without considering the hardware design freedom available in 
many cloud and edge computing applications. For instance, the cloud platforms (e.g. Amazon AWS \cite{amazon2017f1} and Microsoft Azure \cite{Azure}) employ Field Programmable Gate Array (FPGA) for neural network acceleration, while the edge computing platforms typically take the programmable FPGAs \cite{wang2018design,shafiq2017automated} or Application-Specific Integrated Circuit (ASIC) \cite{venkataramani2017scaledeep,whatmough2017dnn}. In addition to neural architecture design, those hardware platforms can also be programmed or even fully customized for the best performance, expanding a \textit{hardware design space}. 

Interestingly, the hardware design space is tightly coupled with the architecture search space, i.e., the best neural architecture depends on the hardware (hardware-aware NAS), and the best hardware depends on the neural architecture. It is therefore best to jointly explore both spaces to push forward the Pareto frontier between hardware efficiency and test accuracy for better design tradeoffs. This can be clearly seen from the example in  Table~\ref{Tab:Evolution}, where three designs on CIFAR-10 and Xilinx XC7Z015 FPGAs are presented: an optimized neural architecture for a fixed FPGA implementation through hardware-aware NAS (design A), the hardware of which is then further optimized through FPGA optimization (design B)~\cite{zhang2016energy}, and a jointly optimized neural architecture and hardware through our co-exploration (design C). From the table, we can see that further optimizing the hardware for the architecture from hardware-aware NAS can lead to $45.45\%$ higher throughput, $38.24\%$ higher energy efficiency with the same accuracy. On the other hand, compared with such a sequential optimization strategy, our co-exploration approach can identify an architecture with higher accuracy and its tailor-made hardware with $16.33\%$ and $28.80\%$ improvements in throughput and energy efficiency, respectively.

\begin{table}
  \centering
  \tabcolsep 7pt
  \renewcommand\arraystretch{1.5}
  \scriptsize
  \caption{On CIFAR-10 and Xilinx XC7Z015 FPGA: Comparisons of three neural  architecture and hardware design pairs in accuracy, throughput, and energy efficiency (E.-E): A) optimal architecture on a fixed hardware implementation through hardware-aware NAS; B) the same architecture but with further FPGA optimization, and C) a jointly optimized neural architecture and FPGA implementation through our co-exploration.}
    \begin{tabular}{ccccc}
    \hline
    \multirow{2}{*}{ID} & \multirow{2}{*}{Approach} & \multirow{2}{*}{Accuracy} & \multicolumn{ 1}{c}{Throughput } & \multicolumn{ 1}{c}{E.-E } \\
    \multicolumn{ 1}{c}{} & \multicolumn{ 1}{c}{} & \multicolumn{ 1}{c}{} & \multicolumn{ 1}{c}{(FPS)} & \multicolumn{ 1}{c}{(GOPS/W)} \\
   \hline
         A & Hardware-Aware NAS &    84.53\% &       16.2 &       0.84 \\

         B & Sequential Optimization &    84.53\% &       29.7 &       1.36 \\

    {\bf C} & {\bf Co-Exploration} & {\bf 85.19\%} & {\bf 35.5} & {\bf 1.91} \\
\hline
    \end{tabular}
\label{Tab:Evolution}
\end{table}






Specifically, our architecture search space and hardware design space co-exploration framework is shown in Figure~\ref{Fig:IntrIll}(b). 
 The proposed co-exploration can be built on any existing NAS framework \cite{zoph2016neural,cai2018proxylessnas,liu2018darts,bender2018understanding} by expanding it to delve into the hardware design space, where a two-level (fast and slow) exploration is iteratively conducted. 
In the fast exploration, the best hardware design is identified for the sampled neural architectures without lengthy training.
The architectures with inferior hardware efficiency will be quickly pruned, which significantly accelerates the search process.
Thereafter, the superior candidates are trained in the slow exploration for controller update using policy gradient reinforcement learning to explore the coupled architecture search space. The optimization objectives in the hardware design space can be varied according to the design specifications, such as area, monetary cost, energy efficiency, reliability, resource utilization, etc. 
 
In order to illustrate our framework, we choose to use FPGA as a vehicle in this paper, as it has gradually become one of the most popular platforms to implement deep neural networks (DNNs) due to its programmability, high performance and energy efficiency, in particular for low-batch inferences \cite{chung2018serving,fowers2018configurable}. Our co-exploration concept and the general framework, however, can also be easily extended to other hardware platforms such as ASICs. 
Since timing performance on a single FPGA is limited by its restricted resource, it is prevalent to organize multiple FPGAs in a pipelined fashion \cite{jiang2018heterogeneous,zhang2019efficient,geng2018framework,geng2018fpdeep} to provide high throughput (frame per second, FPS).
In such a system, the pipeline efficiency is one of the most important metrics needing to be maximized, since it determines the hardware utilization as well as energy efficiency. 
As such, we use accuracy and pipeline efficiency to guide the exploration of the neural architecture space and hardware design space respectively, while satisfying a given throughput specifications (e.g., $\ge$30FPS for the ordinary camera). 
Experimental results show that the co-exploration approach can significantly push forward the Pareto frontier.
On ImageNet, the proposed co-exploration framework can identify architecture and hardware pairs to achieve the same accuracy, 35.42\% higher throughput, and 54.05\% higher energy efficiency with the reduced search time,
compared with the hardware-aware NAS.

\section{Background and Problem Definition}\label{sec:pre}
\subsection{Neural Architecture Search}

Although the research on the automatic prediction of neural network architectures can trace back to the 1980s \cite{schaffer1992combinations}, after deep neural networks have achieved great success in AI domains, there have been growing interests in generating good neural architectures for the interested dataset recently.
With the fact that the architectures are growing deeper, the search space expands exponentially, leading to more difficulties in exploring the search space.
\todo{In the existing work, there are two mainstreams of architecture search:} (1) employing reinforcement learning \cite{zoph2017learning,zoph2016neural,baker2016designing}, (2) applying evolutionary algorithms \cite{real2017large,xie2017genetic,kim2017nemo}.
The basic idea is to iteratively update hyperparameters to generate better ``child networks'' in terms of accuracy.

Figure~\ref{Fig:IntrIll}(a), without the hardware-aware module, illustrates a typically used reinforcement learning based neural architecture search (NAS) \cite{zoph2016neural} framework.
As shown in this figure, the RNN controller in NAS iteratively predicts child networks from architecture search space.
These child networks will be trained on a held-out dataset to obtain its accuracy.
\todo{Then, accuracy will be used as reward to update the RNN controller.}

Existing work has demonstrated that the automatically resulting architectures can achieve close or even higher accuracy to the best human-invented architectures \cite{zoph2017learning,zoph2016neural}.
However, there are two important problems in searching architectures.
First, the search process is inefficient. \cite{zoph2016neural} reported that 20,000 networks were trained across 500 P100 GPUs over 4 days to find the desired network.
Second, since the search process is hardware oblivious, neither the time performance nor the hardware efficiency can be guaranteed.


Recently, hardware-aware NAS \cite{wu2018fbnet,tan2018mnasnet,cai2018proxylessnas} has been proposed to search architectures for a target hardware platform, as shown in Figure \ref{Fig:IntrIll}(a).
\todo{They always assume a fixed hardware design (e.g., mobile chips) and only explore the architecture search space.}
However, the hardware design freedom is commonly available in many cloud and edge computing applications, like FPGA in cloud platforms \cite{amazon2017f1,Azure}  and ASIC in edge computing platforms \cite{venkataramani2017scaledeep,whatmough2017dnn}.
Without the consideration of hardware design space will lead to inferior designs in hardware efficiency, because the hardware design space and the architecture search space are tightly coupled.





\todo{Compared with the existing work, the main contribution of this work is to propose a framework to co-explore the architecture search space and the hardware design space, as shown in Figure \ref{Fig:IntrIll}(b).}
More specifically, this framework determines the best hardware during the search process, which is tailor-made for the candidate architectures.
\todo{In this way, the framework can obtain a set of superior architecture and hardware design pairs on the Pareto frontier in terms of accuracy and hardware efficiency tradeoffs.
In addition, the search time can be significantly reduced, since we can efficiently prune inferior architectures according to multiple design specifications compared with the hardware-aware NAS.}






\subsection{Implementation of DNNs on FPGAs}

This paper will employ FPGA as a vehicle to study how to co-explore neural architectures and hardware designs.
FPGA has demonstrated its excellent ability to achieve high performance and energy efficiency for low-batch real-time inferences \cite{chung2018serving,fowers2018configurable}. 
Hence, a large amount of work is made in implementing neural networks on FPGAs, in which tools are developed to automatically design accelerators on FPGAs for a given network architecture.
In the early stage, research efforts are mainly focusing on designing accelerators on a single FPGA \cite{zhang2015optimizing,shen2017maximizing,zhang2018dnnbuilder,wei2018tgpa}.
\minorR{Authors in \cite{hao2019fpga} target the edge FPGA, Xilinx PYNQ, and demonstrate the advantages of hardware-aware DNN search and update for a single FPGA.}
Most recently, implementations on multiple FPGAs has become the mainstream \cite{geng2018framework,geng2018fpdeep,zhang2016energy,jiang2018heterogeneous,chung2018serving,fowers2018configurable}, since limited resource on a single FPGA becomes the performance bottleneck.

To fully utilize the computation power provided by multiple FPGAs, a typical technique is to implement the neural network on multiple FPGAs in a pipelined fashion \cite{geng2018framework,geng2018fpdeep,zhang2016energy,jiang2018heterogeneous}.
Figure \ref{ProbOverview} demonstrates one such example, in which a 5-layer network is partitioned into 3 pipeline stages, and each pipeline stage is mapped to a certain FPGA in an available pool.
Finally, those FPGAs are connected as a linear array to function in the pipelined fashion.

\subsection{Definitions and Problem Statement}

The goal of the proposed framework is to find both the neural architectures with the highest test accuracy and hardware design with the guaranteed performance (e.g. timing requirement and \todo{hardware efficiency}).
\todo{In this paper, we will employ the conventional convolutional neural network (CNN) based on the multi-FPGA infrastructure as an example to illustrate such a framework, which is the base for other related problems.
In the following, we will first present the relevant definitions.
Then, we will formally define the problem.
Finally, we will discuss the possible extension.}


\begin{figure}[t]
\begin{center}
\centerline{\includegraphics[width=\columnwidth]{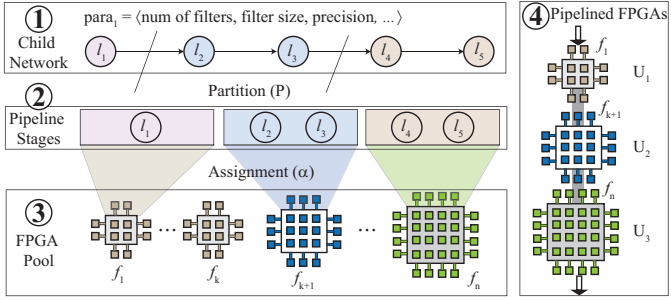}}
\caption{An overview of implementing a child network onto multiple FPGAs to be organized in the pipelined fashion.}
\label{ProbOverview}
\end{center}
\end{figure}

The child network is the bridge between the architecture search space and the hardware design space.
Specifically, in each iteration, the controller RNN will predict child networks from the architecture search space, and then determine their implementations in the hardware design space.
We will introduce the hardware design space as follows.


\textbf{\emph{\raisebox{-1pt}{\large\ding{193}}} Partition Child Network to Pipeline Stages.}
Let $P(C)$ be a set of partitions for the child network $C$.
$P(C)=\{P_1,P_2,\cdots,P_M\}$, where $P_i$ is a nonempty subset of set $L$.
We have the following two properties: (1) $\bigcup_{P_i\in P(C)}=L$; and (2) $\forall P_i, P_j \in P(C)$, if $i\ne j$, then $P_i\cap P_j=\emptyset$.
After the partitioning, each set in $P(C)$ corresponds to a pipeline stage.
For example, in Figure \ref{ProbOverview} \raisebox{-1pt}{\large\ding{193}}, we partition the given child network into 3 pipeline stages, $P_1=\{l_1\}$, $P_2=\{l_2,l_3\}$, and $P_3=\{l_4,l_5\}$.


\textbf{\emph{\raisebox{-1pt}{\large\ding{194}}} Assign Pipeline Stages to FPGAs.}
Then, we can assign each pipeline stage to a specific FPGA in an available FPGA pool, as shown in Figure \ref{ProbOverview} \raisebox{-1pt}{\large\ding{194}}.
An FPGA pool with $n$ FPGAs can be represented by a set $F=\{f_0,f_1,\cdots,f_n\}$.
Each FPGA, $f_i$, has a set of attributes, including memory $mem_i$, DSP slices $dsp_i$, etc.
These attributes will be utilized to model the timing performance for a child network.

We define the assignment function $\alpha$ from the partition set $P(C)$ to FPGA pool $F$.
We have $\alpha(P_i)=f_j$ to indicate the $i^{th}$ pipeline stage $P_i$ is assigned to the $j^{th}$ FPGA $f_j$ to be implemented.
After pipeline stages are assigned to FPGA pool according to $\alpha$, each FPGA will process one or multiple layers.
And all FPGAs work together in the pipelined fashion.

\textbf{\emph{\raisebox{-1pt}{\large\ding{195}}} Pipelined FPGAs.}
The pipelined executions of multiple FPGAs are illustrated in Figure \ref{ProbOverview} \emph{\raisebox{-1pt}{\large\ding{195}}}.
The system will continuously obtain inputs from the dataset with a fixed rate (frame per second), and generate output data from the last pipeline stage.
The input rate of the system reflects the throughput specification $TS$, which implies that the latency of each pipeline stage should be no more than $1/TS$.

The latency of a pipeline stage under an assignment function can be easily captured with a performance model \cite{zhang2015optimizing}.
For FPGA $f_i$, its latency is denoted as $Lat_i$.
After obtaining the latency of each FPGA, we introduce pipeline efficiency, which is composed of the hardware utilization in each pipeline stage (corresponding to an FPGA).
The utilization of FPGA $f_i$ is equal to $Lat_i\times TS$. Higher utilization of an FPGA indicates the less idle time in processing and higher energy efficiency. Therefore, high average utilization of all FPGAs is always desired.


\textbf{Problem Statement.} Based on the above definitions, we formally define the problem of ``hardware/software co-exploration of neural architectures'' as: Given a dataset, a pool of FPGAs $F$, and a throughput specification $TS$, we are going to co-explore architecture search space and hardware design space to find a child network $C$:
\begin{itemize}[noitemsep,topsep=0pt,parsep=0pt,partopsep=0pt]
    \item $para$: parameters of all layers in the child network;
    \item $P$: the partition of layer set $L$ in the child network;
    \item $\alpha$: the assignment of pipeline stages to set $F$;
\end{itemize}
such that the accuracy of child network $C$ is maximized, the pipeline FPGA system can meet the required throughput $TS$, and the average utilization of all FPGAs is maximized.

\todo{
\textbf{Extensions.} The targeting problem is the basis for more general problems.
Therefore, the proposed framework in the next section can be applied to different scenarios with little or no modifications.
In the following, we will discuss different extensions from both hardware and software perspectives.}

\todo{From the hardware perspective, the fundamental problem of mapping child network onto multiple FPGAs is equivalent to that of mapping child network onto multiple processing elements (PEs) in one FPGA, \minorR{where each PE indicates a processor for one data tile (aka. layer processor in \cite{shen2017maximizing})}.
Splitting one FPGA to multiple PEs \cite{shen2017maximizing} is a promising solution when the single FPGA is large enough or the size of neural architecture is relatively small.
In this scenario, a PE can be regarded as an FPGA in the hardware pool in Figure \ref{ProbOverview}.
To apply the proposed technique, we only need to iteratively generate a PE pool (i.e., the number of PEs and the size of each PE) according to the FPGA resource, and conduct co-exploration to identify the best solution for each PE pool.}

\todo{From the software perspective, first, the proposed framework can handle neural networks with residual connections by integrating techniques in \cite{jiang2018design} to partition DAG-based child network; second, it can explore different operations (e.g., group convolutions, depthwise separable convolution, etc.) for each node in a child network by adding one additional parameter in $para_i$ to determine a specific operation for the node.}

\todo{Finally, throughput (frame per second, FPS) in the above problem is set as a constraint. But we can wrap a binary search procedure to maximize throughput together with the pipeline utilization.
Kindly note that by replacing the metrics of FPS to operation per seconds (OPS), the proposed framework can also be applied to optimize other efficiency metrics, like OPS/LUT or OPS/DSP.}

\todo{In the following of this paper, we will focus on determining the best neural architectures and hardware implementations with the conventional CNN structure and multi-FPGA scenario, using the throughput as a constraint and maximizing the hardware utilization.}


\section{HW/SW Co-Exploration Framework}\label{sec:frame}

In this section, we will present the proposed framework. We will use the NAS discussed in \cite{zoph2016neural} as the backbone framework and FPGA as the hardware platform to demonstrate our concept. It can be integrated with any existing NAS techniques \cite{zoph2016neural,cai2018proxylessnas,liu2018darts,bender2018understanding} or extended to incorporate other hardware platforms.  



\subsection{Framework Overview}

\begin{figure}[t]
\begin{center}
\centerline{\includegraphics[width=\columnwidth]{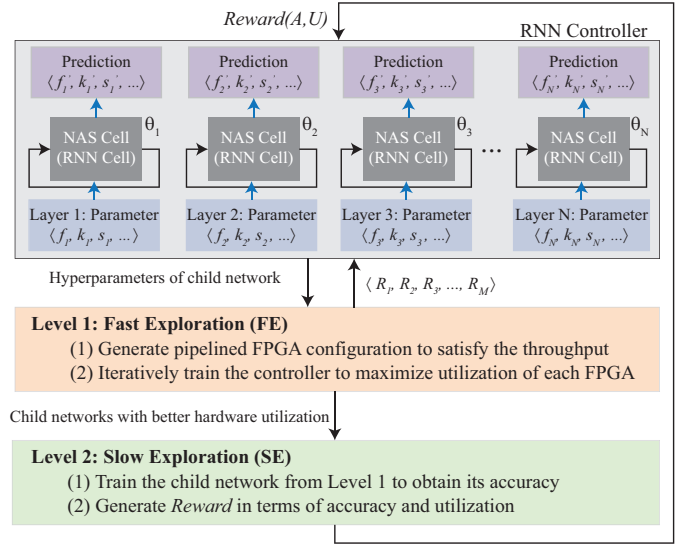}}
\caption{An overview of HW/SW co-exploration framework: The controller contains multiple reconfigurable RNN cells and predicts the hyperparameters in a child network; the fast exploration level prunes child networks with inferior hardware utilization; the slow exploration level updates controller using hardware utilization and accuracy obtained by training child networks.}
\label{overview}
\end{center}
\end{figure}

Figure \ref{overview} shows the HW/SW co-exploration framework.
The framework contains a RNN based controller and two levels of explorations.
Unlike that in \cite{zoph2016neural}, the controller has multiple RNN cells instead of one.
More specifically, each layer in a child network has a corresponding RNN cell.
During the exploration, cells will be reorganized to support different optimization goals.



In the first level, a fast exploration is carried out in four steps: (1) it first predicts an architecture with probability $p$, (2) then, it explores the design space to generate a pipelined FPGA system to meet the throughput requirement, (3) according to the pipeline structure, it then reorganizes RNN cells in the controller, and (4) it updates the controller using reinforcement learning to maximize the pipeline efficiency. This level explores the hardware design space without training child networks, therefore it performs efficiently.



In the second level, we train the child network obtained from the first level on the held-out validation set.
After that, we generate a reward based on both the yielded accuracy and pipeline efficiency, which is used to update the RNN controller.
In case that no child network can meet the required throughput specification in the first level, we generate a negative reward to update the controller.
After this level, the controller will predict a new child network from architecture search space for the fast exploration level.

\todo{
\minorR{The proposed controller integrated with multiple RNNs, operated in two levels of optimizations as shown in Figure \ref{overview}, can make a better tradeoff between efficiency and accuracy.
First, in Level 1, RNNs operate independently to optimize a given architecture for each pipeline stage.
As a result, it can explore the search space more efficiently.
On the other hand, RNNs will work together in Level 2 to determine the backbone architecture and pipeline structure.
Specifically, let $D_i=10^3$ be the size of search space for pipeline stage $p_i$.}
The proposed controller with multiple RNN can optimize each pipeline stage independently, and therefore, the design space is $O(\sum_i\{D_i\})$ (i.e., $O(10^3)$ in the example).
On the contrary, for the controller with only one RNN, it will jointly determine sub-structure for all pipeline stages, leading the search space to be $O(\prod_i{D_i})$ (i.e., $O(10^9)$).
Kindly note that a huge design space will not only significantly prolong the exploration time, but also make it difficult to find the best solution.
The advantages of the proposed framework in both efficiency and effectiveness will be verified in the experimental results.
}









\subsection{Fast Exploration for High Resource Utilization}\label{sec:fast}

In the first level, namely Fast Exploration (FE), the objective is to maximize pipeline efficiency under the throughput specification $TS$.
FE takes three types of inputs: (1) a set of available FPGAs $F$, (2) hyperparameters of a child network $H$, (3) a throughput specification $TS$.
It will generate a new child network, whose throughput at inference phase can meet $TS$ using a subset of FPGAs in $F$.
In addition, the average hardware utilization of FPGAs can be maximized.
In FE, there are two challenges needing to be addressed: first, how to partition a given child network and assign each partition to a specific FPGA (Partition and Assignment); second, how to reorganize the RNN cells in the controller and then update them to generate child networks with higher pipeline efficiency (Reorganize and Update Controller).

\begin{figure}[t]
\begin{center}
\centerline{\includegraphics[width=\columnwidth]{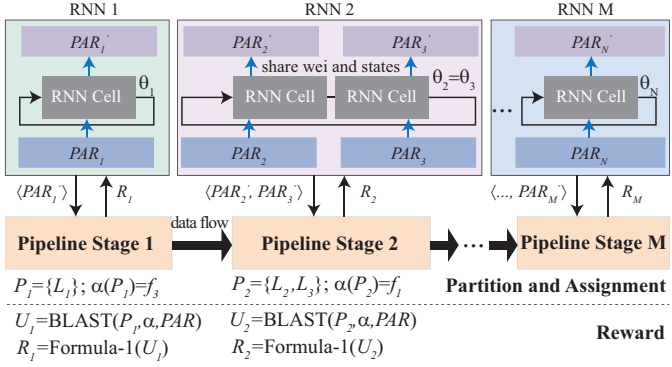}}
\caption{Fast Exploration (FE): organize RNN cells in the controller according to the partition for pipeline stages; independently update multiple RNNs in the controller to predict parameters of layers assigned to each pipeline stage.}
\label{fig:levele1}
\end{center}
\end{figure}

\textbf{Partition and Assignment}. In the search process, a number of candidate child networks need to go through the partition and assignment process.
Consequently, an efficient automatic tool should be employed to avoid performance degradation on search process.
In this paper, we employ the BLAST algorithm in \cite{jiang2018heterogeneous}.
BLAST takes child network $H$, FPGAs $F$, the throughput specification $TS$, and the attributes of each FPGA as inputs.
It outputs a serial of FPGAs, each of which will implement one or multiple layers in a pipeline stage.
The resultant system will satisfy $TS$ with the maximum pipeline efficiency.
As shown in Figure \ref{fig:levele1}, layers in a child network are divided into $M$ partitions, and each partition is assigned to one specific type of FPGA under function $\alpha$.









\textbf{Reorganize and Update Controller}. According to the generated pipeline structure, we then reorganize the controller and iteratively update the controller to generate child networks with higher hardware utilization.
Our goal is to maximize the average hardware utilization, which is equivalent to maximize the utilization of each hardware.
However, the design space of maximizing the average hardware utilization is exponentially larger than that of maximizing the utilization of each hardware.
To efficiently explore the design space, we choose to maximize the hardware utilization of different pipeline stage independently.
Therefore, we reorganize RNN cells in the controller according to the determined pipeline structure.
More specifically, for multiple layers in one pipeline stage, their corresponding RNN cells will be configured to form one RNN and their weights and states are shared (e.g., RNN 2 in Figure \ref{fig:levele1}).
In consequence, there will be $N$ RNNs for $N$ pipeline stages. In this way, each RNN can be trained to maximize the hardware utilization for each FPGA pipeline stage.







After we form the RNNs, we apply reinforcement learning to update the parameters in those $N$ RNNs, and use these RNNs to predict the hyperparameters of child networks.
In each iteration, we will predict $T$ child networks, which can be viewed as a list of actions $a_{1:T}$.
Correspondingly, notation $a_{1:T}^{i}$ represents the hyperparameters of the $i^{th}$ pipeline stage in these child networks.
For each child network predicted by the controller, we can obtain the utilization of the $i^{th}$ pipeline stage (corresponding to one FPGA) using BLAST, denoted as $U_i$.
Then, for RNN $i$, we utilize $U_i$ to generate a reward $R_i$ to update its parameters $\theta_i$.
The reward $R_i$ can be calculated using the following formula.
\begin{equation}
R_{i} = \left\{ \begin{matrix}
U_{i} & U_{i} \leq 1 \\
1 - U_{i} & {1 < U}_{i} \leq 2 \\
 - 1 & U_{i} > 2 \\
\end{matrix} \right.\
\end{equation}
where $U_i>1$ indicates that the required throughput cannot be satisfied, and we give the negative reward.
For each RNN, our objective is to maximize the expected reward for actions from time 1 to $T$, represented by $J(\theta_i)=E_{P(a_{1:T}^i;\theta_i)}[R_i]$.
Since the reward is non-differentiable, we apply the policy of gradient method to update $\theta_i$.
Specifically, the method of REINFORCE rule \cite{williams1992simple} has been employed as in \cite{zoph2016neural,cai2018proxylessnas}.

\subsection{Slow Exploration for High Accuracy}

\begin{figure}[t]
\begin{center}
\centerline{\includegraphics[width=\columnwidth]{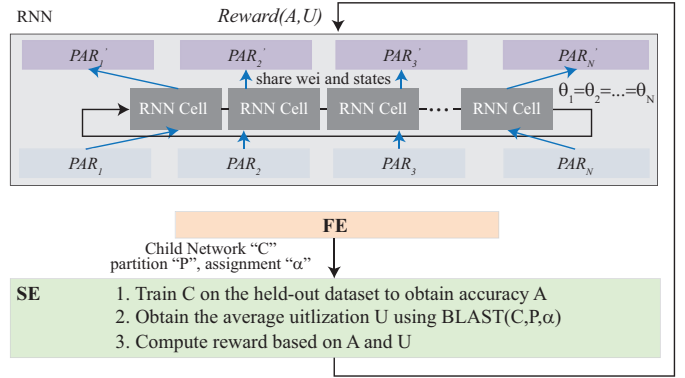}}
\caption{Slow Exploration (SE): configure RNN cells in the controller to be one RNN; generate reward based on accuracy and pipeline efficiency to update the controller RNN.}
\label{fig:levele2}
\end{center}
\end{figure}

After obtaining a child network meeting the timing specification through the fast exploration level, we now move to the second level. 
In this level, we aim to update the controller RNN to generate new child networks with higher accuracy and pipeline efficiency.
We will train the child network on the held-out validate set, and therefore the exploration speed is much slower than that of the first one. We call it Slow Exploration (SE).


As shown in Figure \ref{fig:levele2}, SE takes the generated child network, the partition and the assignment from FE as the inputs.
The child network is first trained to obtain accuracy $A$.
Then, the average pipeline efficiency $U$ of the child network under the partition and assignment will be calculated.
Finally, we compute the reward to update the controller using the following formula.
\begin{equation}\label{equ:rwd}
Reward(A,U) = \beta\times A + (1-\beta)\times U
\end{equation}
where $\beta$ is an adjustment parameter, which reflects the bias on test accuracy and hardware utilization. The value of $\beta$ ranges from 0 to 1.
We will discuss how to scale $\beta$ in Section \ref{Sec:Res}.
After that, we update the controller using the reward by applying the policy gradient reinforcement learning, which is the same as that in FE level.
As shown in Figure \ref{fig:levele2}, all RNN cells share the same weights and states in this level, since we have only one reward.

\subsection{Interface between Fast-Slow Explorations}

Before updating the RNN cells in the controller in the fast exploration level, we take a snapshot $Snap$ of all RNN cells.
During the fast exploration level, we obtain the hardware design (i.e., pipeline configuration) for the input child network.
Based on the determined pipeline structure, RNN cells are reorganized as introduced in Section \ref{sec:fast}.
And reorganized cells will be trained to generate better child networks for the previously obtained hardware design (i.e., pipeline configuration). Finally, a child network with maximum hardware efficiency on the determined pipeline will be sent to the slow exploration level.

After entering the slow exploration level, the RNN cells in the controller will be recovered using the previously saved snapshot $Snap$. Then, SE will train the child network to obtain the accuracy, which will be used to calculate the reward. Using this reward, we will update the recovered RNN. Then, the updated RNN will be used to generate new child networks for the next iteration. In this way, the SE process will always keep improving the RNN accuracy while the FE process will always generate the best hardware design for each iteration. 


%






\section{Experiments}\label{sec:exp}




\textbf{Datasets:} We use CIFAR-10 and ImageNet datasets to study the efficacy of our approach and compare it with the state-of-the-art.
During the exploration of child networks, we only use the training images in these datasets, while the test images are used to test the accuracy of the resultant architectures.
To evaluate the accuracy in the search process, we randomly select 10\% of the samples from the training set as a validation set.
All the images undergo the data preprocessing and augmentation procedure, including whitening, upsampling, random cropping, and random horizontal flip, which are common among the related work.


\textbf{Architecture Search Space:} For CIFAR-10, we use convolutional architectures as the backbone.
For every convolutional layer, we first determine the filter size in [24,36,48,64], the kernel size in [1,3,5,7], and the strides.
Two sets of experiments are carried out to determine the strides: (1) by exploring the child networks with a fixed stride of 1; (2) by allowing the controller to predict the strides in [1,2].
After each layer, the rectified linear units \cite{nair2010rectified} and the batch normalization \cite{ioffe2015batch} are appended.

For ImageNet, the architecture repeats mobile inverted bottleneck convolution layers instead of ordinary convolutional ones, same as that in \cite{cai2018proxylessnas}.
The controller explores the architectures with various kernel sizes [3,5,7], strides [1,2] and expansion ratios [3,6].


\textbf{Hardware Design Space:} The hardware design space is composed of up to three Xilinx FPGAs (XC7Z015), each of which contains 74K logic cells, 4.9Mb on-chip memory, and 150 DSP Slices.
One reason for our selection is that such an FPGA provides high speed serial communication (up to 16.8Gbps of bandwidth), so that a high speed hardware pipeline can be formed by multiple FPGAs. 
In the implementation, the child network is partitioned into pipeline stages, and each stage is mapped to one FPGA.
Kindly note that our hardware exploration may not end up using all three FPGAs; it is possible to use fewer for higher hardware efficiency.

In the experiments, we use pipeline efficiency as the metrics to measure the hardware efficiency.
As stated in Section \ref{sec:Intro}, the pipeline efficiency is one of the most important metrics, since it is related to the hardware utilization, energy efficiency, and timing performance.
Then, the timing specifications are set according to the desired processing speed of the data at the inference phase, which are commonly decided by the data collector (e.g., camera).
For CIFAR-10, we set the throughput specification to 35FPS, which can satisfy most cameras; whereas for ImageNet, due to the more complicated architectures and the limited resource, we set the specification to 10FPS.
\todo{Finally, for both data and weights, we apply the commonly used 16-bit fixed point data, as that in \cite{ma2019performance,jiang2018heterogeneous,zhang2015optimizing,shen2017maximizing}.
}

\textbf{Training Details: }
For CIFAR-10, the training settings for both the RNN controller and the child networks are the same as \cite{zoph2016neural}.
For the controller RNN, in both slow and fast explorations, it is trained by using the calculated rewards with the ADAM optimizer \cite{kingma2014adam} with a learning rate of 0.0006.
Parameter $\beta$ in Formula \ref{equ:rwd} is set to 0.5 to equally optimize test accuracy and pipeline efficiency.
For the child networks, we apply Momentum Optimizer with a learning rate of 0.1, weight decay of  $10^{-4}$. and momentum of 0.9.
Each child network is trained for 50 epochs.

For ImageNet, we build the distributed GPU training environment on top of Uber Horovod \cite{sergeev2018horovod}.
Training settings are similar to those for CIFAR-10, with the 
exceptions that we set the initial learning rate to $0.0125$, decay 10$\times$ at selected epochs, and for the Momentum Optimizer the weight decay is $5\times 10^{-5}$ and the momentum is 0.9.

\section{Results} \label{Sec:Res}


\todo{This section will report comparison results in four sets of experiments: (1) we compare the proposed framework with different configurations; (2) we compare the proposed framework with the existing NAS frameworks; (3) we compare the identified architectures with the existing ones; (4) we show the design space exploration in terms of model size and hardware efficiency to demonstrate the importance of hardware/software co-exploration.}

\begin{figure}[t]
\begin{center}
\centerline{\includegraphics[width=3.2in]{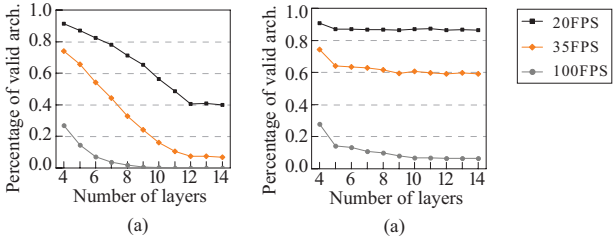}}
\caption{Percentages of valid architectures for different timing specifications: (a) fixed stride of 1; (b) predictable strides.}
\label{fig:expSetting}
\end{center}
\end{figure}

\todo{
\subsection{Comparison Results with Different Configurations}
Before reporting the results, we first introduce the setting for the proposed framework, namely ``Co-Exploration''.
First, the search spaces and training settings can be found in Section \ref{sec:exp}.
Second, the controller will iteratively search child networks for 10,000 episodes through the 2-level exploration.
Third, in each episode, the slow exploration phase will obtain accuracy of 16 child networks (train from scratch if one has never been trained or obtain accuracy from a history table); these child networks are identified by the fast exploration phase, where 100 trails will be taken for each child network to optimize the hardware efficiency.
Since the proposed framework has multiple optimization goals on both software (e.g., accuracy) and hardware (e.g., pipeline efficiency), we record a set of superior architecture and hardware design pairs during the exploration, which forms the Pareto frontier. 
On the frontier, we denote the solution with the maximum accuracy as ``OptSW'' and the solution with the maximum pipeline efficiency as ``OptHW''. 
}

\textbf{Impact of Timing Specifications:} Figure \ref{fig:expSetting} reports the impact of timing specifications for the Co-Exploration framework.
We randomly sample 10,000 architectures for the layer size ranged from 4 to 14, and obtain the percentage of valid architectures that can meet the timing specification on the CIFAR-10 dataset.
In Figure \ref{fig:expSetting}, it is obvious that if the constraint is tight (e.g., FPS=100), only a few architectures can satisfy the specification, indicating that the number of architectures with high accuracy is reduced compared with the one without timing constraints.
In this case, we can scale up the parameter $\beta$ in Formula \ref{equ:rwd} to pursue higher accuracy.
On the other hand, if the constraint is loose (e.g., FPS=20), there are a large number of valid architectures.
Correspondingly, we can scale down $\beta$ to find more hardware efficient designs with high accuracy.




\begin{table}
  \centering
  \tabcolsep 6pt
  \renewcommand\arraystretch{1.5}
  \scriptsize
  \caption{Co-Exploration with predictable stride performs better than that with fixed stride under 35FPS timing specification.}
\begin{tabular}{cccc}
\hline
  Models &      Depth &   Accuracy & Pipeline Eff. \\
\hline  
Co-Exploration fixed stride (OptSW) &         13 &    81.50\% &    91.92\% \\
Co-Exploration fixed stride (OptHW) &         10 &    78.57\% &    98.56\% \\
\hline
Co-Exploration pred. stride (OptSW) &         14 & {\bf 85.19\%} &    92.15\% \\
Co-Exploration pred. stride (OptHW) &          6 &    80.18\% & {\bf 99.69\%} \\
\hline
\end{tabular}
\label{Tab:CIFARResults}
\end{table}


\begin{table*}
  \centering
  \tabcolsep 10.5pt
  \renewcommand\arraystretch{1.5}
  \scriptsize
  \caption{Comparison among Co-Exploration, Hardware-Aware NAS and Sequential Optimization on CIFAR-10 and ImageNet datasets.}
\begin{tabular}{cccccccccc}
\hline
\multirow{2}{*}{Dataset}  &    \multirow{2}{*}{Models} &      \multirow{2}{*}{Depth} & \multirow{2}{*}{Parameters} & Accuracy & Accuracy & \multirow{2}{*}{Pipeline Eff.} &        \multirow{2}{*}{FPS} & Energy Eff.  \\
    &       &   &            &       (Top1)            &     (Top5)         &  &               &                GOPS/W         \\
\hline
\multirow{4}{*}{CIFAR-10} & Hardware-Aware NAS &         13 & 0.53M &    84.53\% &    - &  73.27\% &  16.2 &               0.84 \\

& Sequential Optimization &         13 & 0.53M &   84.53\% &    - &   92.20\% &   29.7 &               1.36  \\

\cline{2-9}
& Co-Exploration (OptHW) &         10 & 0.29M &    80.18\% &    - & {\bf 99.69\%} &  {\bf 35.5} &      {\bf 2.55}  \\

& Co-Exploration (OptSW) &         14 & 0.61M & {\bf 85.19\%} & - &   92.15\% & {\bf 35.5}   &         1.91  \\
\hline

\hline




\multirow{4}{*}{ImageNet} & Hardware-Aware NAS  &         15 & 0.44M &     68.40\% &    89.84\% &  81.07\% &   6.8 &             0.34  \\

& Sequential Optimization &         15 &   0.44M &   68.40\% &    89.84\% &    86.75\% & 10.4 &              0.46  \\

\cline{2-9}
& Co-Exploration (OptHW) &         17 &   0.54M  &  68.00\% &    89.60\% &   {\bf 96.15\%} & {\bf 12.1}&    {\bf 1.01} \\

& Co-Exploration (OptSW) &         15 & 0.48M  &  {\bf 70.24\%} & {\bf 90.53\%} &  93.89\% & {10.5}  &        0.74  \\
\hline

\end{tabular}
\label{Tab:ImageResults}
\end{table*}

\begin{figure}[t]
\begin{center}
\centerline{\includegraphics[width=\columnwidth]{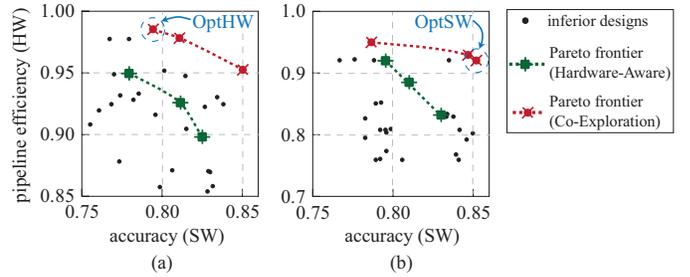}}
\caption{Pareto frontiers between accuracy and pipeline efficiency for Hardware-Aware NAS and Co-Exploration, both of which are designed under the timing specification of 35FPS: (a) designs with 2 FPGAs; (b) designs with 3 FPGAs.}
\label{fig:expSpace}
\end{center}
\end{figure}

\textbf{Comparison between Fixed Stride and Predictable Stride:} 
Table \ref{Tab:CIFARResults} reports the comparison between the exploration with the fixed stride and that with the predictable stride on CIFAR-10\todo{\footnote{\todo{Models accessed at: https://github.com/PITT-JZ-COOP/Co-Explore-NAS}}}.
In the table, column ``depth'' indicates the number of layers in the resulting architecture.
As shown in this table, for the exploration with the fixed stride, OptSW achieves 2.93\% higher accuracy but 6.64\% loss in pipeline efficiency than OptHW.
These figures are 5.01\% and 7.54\% for the exploration with the predictable strides.
In addition, it is obvious that compared with fixed stride, the stride prediction can help controller to find better results in both accuracy and pipeline efficiency.
As such, in the following experiments we will use predictable stride as the default setting for Co-Exploration.

\todo{
\subsection{Comparison Results with the Existing NAS Frameworks}
Next, we compare the proposed Co-Exploration framework with the existing NAS frameworks.
To be fair, we use the same setting as the Co-Exploration: exploring 10,000 episodes and getting accuracy of 16 child networks in each episode.
Because the existing Hardware-Aware NAS frameworks \cite{wu2018fbnet,cai2018proxylessnas,tan2018mnasnet} target  fixed hardware (e.g., GPU) instead of programmable FPGAs, and they use various settings; for fair evaluation, we use the NAS discussed in  \cite{zoph2016neural} as the backbone to implement a Hardware-Aware NAS for FPGA with the same search spaces and training settings as described in Section \ref{sec:exp}.
Unlike the Co-Exploration framework, the Hardware-Aware NAS assumes fixed accelerator designs (i.e., optimization parameters) in FPGAs.
As shown in Figure \ref{Fig:IntrIll}(a), in the search loop, the controller will first predict a neural architecture; second, the framework tests the hardware efficiency of the predicted architecture on FPGAs; third, it trains architecture to get its accuracy; finally, it utilizes hardware efficiency and accuracy to update the controller.
This framework is denoted as Hardware-Aware NAS in the results.}

\todo{In addition, for the final architectures obtained by 
the Hardware-Aware NAS, we further optimize their hardware implementation to achieve a better design in terms of hardware efficiency. 
Such a heuristic approach is denoted as ``Sequential Optimization'' in the results.
}




\textbf{Impact of Different Exploration Frameworks on Pareto Frontier:} Figure \ref{fig:expSpace} reports the design space exploration assuming the hardware design space contains up to (a) two FPGAs or (b) three FPGAs. 
The x-axis and y-axis represent the accuracy and pipeline efficiency, respectively. 
For clear demonstration, we only include the architectures whose pipeline efficiency is no less than 85\% for two FPGAs in Figure \ref{fig:expSpace}(a) and no less than 75\% for three FPGAs in Figure \ref{fig:expSpace}(b). 
In these figures, the circled design points correspond to those in Table \ref{Tab:CIFARResults}. The red lines represent the Pareto frontiers explored by Co-Exploration. The green lines, on the other hand, represent the frontier obtained by Hardware-Aware NAS (by examining the top  architectures identified). These figures clearly show that by exploring hardware design space, our Co-Exploration can significantly push forward the Pareto frontiers in the accuracy and efficiency tradeoffs. It effectively identifies better designs not available through architecture search space only, i.e., those between the two frontiers.

Comparing the two exploration results in Figure \ref{fig:expSpace}(a) and (b), we can also see that the solution with the highest pipeline efficiency is located in Figure \ref{fig:expSpace}(a), while the one with the highest accuracy is located in Figure \ref{fig:expSpace}(b).
In general, we can always observe that the average accuracy on three FPGAs is higher than that on two FPGAs, yet the pipeline efficiency is lower.
This is because more FPGAs can accommodate deeper architecture in layers for higher accuracy. On the other hand, more layers will easily result in unbalanced pipeline stages, which in turn reduces the pipeline efficiency.

\textbf{Comparison between Co-Exploration and Existing Frameworks:} 
Table \ref{Tab:ImageResults} reports the comparison results on accuracy, pipeline efficiency, throughput and energy efficiency on CIFAR-10 and ImageNet.
All the architectures identified have fewer than 1M parameters mainly due to the hardware capacity.
This inevitably leads to accuracy loss; however, as we can see, the architecture explored by OptSW can still achieve 85.19\% test accuracy on CIFAR-10, and 70.24\% top-1 accuracy on ImageNet.
These results demonstrate the effectiveness of the Co-Exploration approach in resource limited scenarios.
In addition, OptSW outperforms Hardware-Aware NAS by achieving $54.37\%$ and $35.24\%$ higher throughput, and $56.02\%$ and $54.05\%$ higher energy efficiency on CIFAR-10 and ImageNet, respectively. 
Compared with Sequential Optimization, OptSW achieves $16.34\%$ and $28.79\%$ improvements on CIFAR-10 in throughput and energy efficiency, respectively; and on ImageNet, it can also slightly improve throughput, and achieve $37.84\%$ improvements on energy efficiency.


\begin{table}
  \centering
  \tabcolsep 2pt
  \renewcommand\arraystretch{1.5}
  \scriptsize
  \caption{Co-Exploration uses much fewer GPU hours than that of Hardware-Aware NAS, benefiting from the early-stage pruning.}
\begin{tabular}{ccccc}
\hline
   Dataset &   Approach & Arch for Training &  GPU Hours &      Impr. \\
\hline
\multirow{2}{*}{CIFAR-10} & Hardware-Aware NAS &    108,000 &     16,586 &          1 \\

\multicolumn{ 1}{c}{} & Co-Exploration &        308 &        102+1.9=103.9 &        159$\times$ \\
\hline
\multirow{2}{*}{ImageNet} & Hardware-Aware NAS &      7,263 &     36,315 &          1 \\

\multicolumn{ 1}{c}{} & Co-Exploration &         53 & 256+1.8=266.8 &        136$\times$ \\
\hline
\end{tabular}  
\label{Tab:GPUHour}
\end{table}

\begin{table}
  \centering
  \tabcolsep 5pt
  \renewcommand\arraystretch{1.5}
  \scriptsize
  \caption{\todo{Comparison with the existing architectures on ImageNet with the timing specification of 10FPS.}}
\begin{tabular}{cccccc}
\hline
    \multirow{2}{*}{Models} &      \multirow{2}{*}{Depth} & Accuracy  & Accuracy  &        \multirow{2}{*}{FPS} & \multirow{2}{*}{Energy Eff.}  \\
    & & (Top-1) & (Top-5) & &  \\
\hline
MobileNetV2 \cite{sandler2018mobilenetv2} &         18 &    71.80\% &    91.00\% &        4.5 &       0.47    \\

ProxylessNet \cite{cai2018proxylessnas} &         21 &    74.60\% &    92.50\% &        3.1 &       0.41  \\

\hline

Co-Exploration (OptHW) &         17 &    68.14\% &    89.60\% & {\bf 12.1} & {\bf 1.01}  \\

Co-Exploration (OptSW) &         15 &    \textbf{70.24\%} &    \textbf{90.53\%} &       10.5 & 0.74  \\
\hline
\end{tabular}
\label{Tab:NewImageResults}
\end{table}

Finally, Table \ref{Tab:GPUHour} reports the comparison results on normalized search time between the Hardware-Aware NAS and the Co-Exploration.
Results in this table show that the Co-Exploration can significantly accelerate the search process, achieving $159\times$ and $136\times$ fewer GPU hours on CIFAR-10 and ImageNet, respectively.
The speedup is achieved from the efficient early-stage pruning in the fast exploration level.
\todo{As discussed in Section \ref{sec:frame}-A, compared with the conventional Hardware-Aware NAS with a single RNN in the controller, the proposed Co-Exploration framework with multiple RNNs can dramatically shrink the design space from $O(\prod_i{D_i})$ to $O(\sum_i{D_i})$, where $D_i$ is the size of design space for the $i^{th}$ pipeline stage.
Since the number of architecture to be trained is proportional to the size of design space, from column ``Arch for Training'' in Table \ref{Tab:GPUHour}, we can see that Co-Exploration trains much fewer architectures compared with the Hardware-Aware NAS.
Therefore, our Co-Exploration achieves significant speedup over the Hardware-Aware NAS.
From the table, we have another observation that the training process takes much longer time than the hardware exploration process, where the hardware exploration only occupies less than 1\% GPU hours in the whole search process (1.9 GPU hours for CIFAR-10 and 1.8 GPU hours for ImageNet).
}


\todo{
\subsection{Comparison Results with the Existing Architectures}
In this subsection, we compare the neural architectures identified by the proposed Co-Exploration framework with the existing architectures: ProxylessNet \cite{cai2018proxylessnas} and MobileNetV2 \cite{sandler2018mobilenetv2}.
We set the throughput constraint as 10FPS for Co-Exploration framework as a baseline.
To obtain the hardware efficiency (throughput, energy efficiency, etc.) of these architectures, \minorR{we apply the BLAST approach \cite{jiang2018heterogeneous} to partition them onto multiple FPGAs. For the fair of comparison, all models involve 3 FPGAs.} 
}

\todo{
Table \ref{Tab:NewImageResults} reports the results.
As we can see, both MobileNetV2 and ProxylessNet cannot meet the timing specification of 10 FPS, while ours can.
In comparison with the manually designed MobileNetV2 \cite{sandler2018mobilenetv2}, OptSW with top-5 accuracy loss of 0.47\% can achieve $2.33\times$ and $1.57\times$ improvement on throughput and energy efficiency, respectively.
On the other hand, in comparison with ProxylessNet \cite{cai2018proxylessnas}, whose throughput is $3\times$ lower than the specifications, OptSW can find architectures that meet the specs with 90.53\% top-5 accuracy against 92.50\% from ProxylessNet.
Results show that the proposed framework can make a better tradeoff between hardware efficiency and architecture accuracy.
In addition, it can guarantee that the final architecture identified can meet the timing specification, which is important in real-time AI systems.
}

\begin{figure}[t]
\begin{center}
\centerline{\includegraphics[width=3.4in]{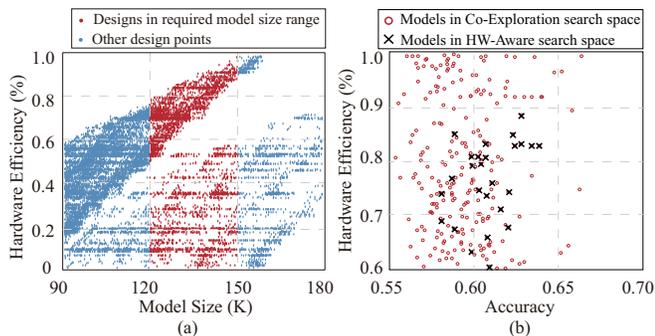}}
\vspace{-10pt}
\caption{\minorR{Design space of architectures with the depth of 4: (a) model size v.s. hardware efficiency; (b) accuracy v.s. hardware efficiency using co-exploration and hardware-aware NAS approaches.}}
\label{fig:DSE}
\end{center}
\end{figure}

\todo{
\subsection{Importance of Co-Exploration}
Finally, we show the importance of co-exploration on NAS and hardware design spaces, instead of (1) using a heuristic on restricting the size of models for only NAS exploration\minorR{, or (2) applying hardware-aware NAS exploration.}
Figure \ref{fig:DSE} shows the results of the design space exploration of architectures with 4 layers.
}

In Figure \ref{fig:DSE}(a), the x-axis and y-axis represent the model size and the hardware efficiency (i.e., pipeline efficiency).
Each point in this figure is a design, which is optimized using the algorithm in \cite{jiang2018heterogeneous}.
We have marked the design points whose model size ranges from 120K to 150K.
From this figure, we can see that, for the designs whose model size ranges from 120K to 150K, the optimized hardware efficiency ranges from 1.29\% to 98.35\%.
Moreover, for a much narrower range from 149K to 150K, the efficiency still ranges from 7.02\% to 98.35\%.
All the above results reflect that we cannot guarantee the hardware efficiency by restricting the model size only.
This is mainly because there are a large number of designs with similar model size, but their structures are quite different, leading to different hardware efficiency.
Therefore, it verifies the neural architecture search space and hardware design space are tightly coupled and emphasizes the importance of conducting hardware and software co-exploration.

\minorR{
In Figure \ref{fig:DSE}(b), we unveil the fundamental difference between co-exploration and hardware-aware architecture search.
In this figure, the black crosses and red circles  represent the valid design points in HW-aware NAS and co-exploration search spaces, respectively.
We can observe that the HW-aware NAS has a much narrower search space than the proposed co-exploration approach.
Basically, HW-aware NAS will prune the architectures with high accuracy but fail to meet hardware specifications on fixed hardware design.
However, by opening the hardware design space, it is possible to find a tailor-made hardware design for the pruned architectures to make them meet the hardware specifications.
Therefore, compared with the HW-aware NAS, the co-exploration approach enlarges the search space. 
As a result, it can make better tradeoffs between accuracy and hardware efficiency.
}

\section{\todo{Conclusion and Future Work}}\label{sec:con}
We proposed the co-exploration framework to open up the hardware design freedom in neural architecture search.
This is driven by the trend that the hardware platform can be programmed or even fully customized for the best performance in cloud and edge computing applications.
This paper took the FPGA as a vehicle to show that through jointly exploring architecture search space and hardware design space, the design Pareto frontier on accuracy and hardware efficiency tradeoffs can be significantly pushed forward.

\todo{The framework proposed in this paper will be the base for neural architecture and hardware co-exploration.
Based on the proposed co-exploration framework, we list two promising future directions as follows.
First, mixed-precision was recently proposed \cite{wang2019haq} for a fixed architecture; in the future, we plan to further co-explore neural architectures, quantizations and hardware designs.
Second, innovations on computing architecture achieves great success for executing inference phase of neural networks \cite{chen201865nm}, we plan to apply the proposed framework to co-explore neural architectures with the novel computing architectures (e.g., computing-in-memory).}







\bibliography{example_paper}
\bibliographystyle{IEEEtran}

\end{document}